\documentclass[runningheads]{llncs}
\usepackage[T1]{fontenc}
\usepackage{graphicx}
\usepackage{comment}
\usepackage{multirow}
\usepackage{array}
\usepackage{subcaption}
\usepackage{url}
\usepackage{xcolor}

\begin{document}
\title{In-domain SSL pre-training and streaming ASR:\\
Application to Air Traffic Control Communications}
\titlerunning{In-domain SSL pre-training and streaming ASR}
%

\author{Jarod Duret\inst{1} \and
Salima Mdhaffar\inst{1} \and
Gaëlle Laperrière\inst{1} \and
Ryan Whetten\inst{1} \and \\
Audrey Galametz\inst{2} \and
Catherine Kobus\inst{2} \and
Marion-Cécile Martin\inst{2} \and
Jo Oleiwan\inst{2} \and
Yannick Estève\inst{1}}
\authorrunning{J. Duret et al.}
%
\institute{{Avignon Université - LIA, France}\\
 \and
AIRBUS}
\maketitle              
\begin{abstract}
In this study, we investigate the benefits of domain-specific self-supervised pre-training for both offline and streaming ASR in Air Traffic Control (ATC) environments. We train BEST-RQ models on 4.5k hours of unlabeled ATC data, then fine-tune on a smaller supervised ATC set. To enable real-time processing, we propose using chunked attention and dynamic convolutions, ensuring low-latency inference. We compare these in-domain SSL models against state-of-the-art, general-purpose speech encoders such as w2v-BERT 2.0 and HuBERT. Results show that domain-adapted pre-training substantially improves performance on standard ATC benchmarks, significantly reducing word error rates when compared to models trained on broad speech corpora. Furthermore, the proposed streaming approach further improves word error rate under tighter latency constraints, making it particularly suitable for safety-critical aviation applications. These findings highlight that specializing SSL representations for ATC data is a practical path toward more accurate and efficient ASR systems in real-world operational settings.

\end{abstract}

\newcommand{\blue}[1]{\textcolor{black}{#1}}
\newcommand{\offlinemodel}[1]{BRQ-ATCO2$_{#1}$}
\newcommand{\streamingmodel}[1]{Stream-ATCO2$_{#1}$}

\section{First Introduction}
Automatic speech recognition (ASR) has become an essential technology in various fields, including aviation, where precise and real-time transcription of spoken communication could become crucial~\cite{helmke2016reducing}. 
Air Traffic Control (ATC) communications represent a particularly challenging domain for ASR due to their constrained but highly specialized vocabulary, strict grammar structures, and wide range of speaker accents. 
These factors, combined with the presence of background noise, make ATC a specific and demanding application for speech recognition systems, especially when real-time processing is targeted.
While SSL-pre-trained models such as wav2vec 2.0~\cite{baevski2020wav2vec} and HuBERT~\cite{hsu2021hubert} have demonstrated strong performance across various ASR benchmarks, their adaptation to highly specialized domains like ATC presents unique challenges that require further investigation~\cite{zuluaga2023does}. 
A key consideration for existing SSL models is that they are often pre-trained on diverse but predominantly general-purpose speech data, which may not fully align with the linguistic and acoustic characteristics of specialized domains like ATC.

In this work, we explore the impact of in-domain SSL pre-training for offline and streaming ASR dedicated to ATC communications. 
To evaluate our approach, we conduct experiments on both proprietary and publicly available ATC datasets.
This paper is organized as follows: first, we discuss related work and the characteristics of ATC communications and datasets. 
Next, we describe the pre-training process of our SSL models and compare their performance for offline ASR. We then introduce a method for pre-training a model for streaming ASR and present experimental results.

\section{Related work}

The application of self-supervised learning (SSL) models to ATC speech processing has gained significant attention in recent years. 
In~\cite{zuluaga2023does}, the authors investigated the suitability of pre-trained SSL wav2vec~2.0 models, for transcribing ATC speech and detecting key information, demonstrating that these models can benefit from domain adaptation techniques to improve recognition performance in this specialized setting. 
In addition to improving accuracy, recent research has also focused on optimizing SSL models for real-time streaming applications~\cite{kanagawa2024knowledge}, which is crucial for ATC scenarios where latency should be minimized for grounded scenarios. 
Another key development in SSL-based streaming ASR is the BEST-RQ model~\cite{chiu2022self}, which introduces quantization techniques for improved representation learning. 
Originally designed to enhance efficiently self-supervised learning with discrete latent representations, BEST-RQ has also been explored for its potential in streaming speech recognition~\cite{chiu2022self}. 

\section{ATC datasets used in this study}
\label{sec:datasets}
The Airbus-ATC corpus is the dataset released for the Airbus ATC Speech Recognition 2018 Challenge~\cite{pellegrini2019airbus}.
A software-defined radio receiver connected to an aeronautical antenna and set to capture local airport ATC communications has been used to record the audio at 16kHz~\cite{delpech-etal-2018-real}. Since the collected audios originate from French airports, the French accent is predominant. 
The Airbus-ATC corpus contains approximately 50 hours of recorded communications gathered from multiple French airports, split into 3 datasets: about 40h for supervised training, 5h for validation, and 5h for evaluation.

The ATCO2 corpus~\cite{zuluaga2022atco2} contains also real-world ATC voice recordings. 
It brings together thousands of unlabelled hours of communications between air-traffic controllers and pilots, drawn from publicly available sources such as LiveATC, and a small part (1h for the free version, 4h for the purchased version) of manually transcribed speech recording.
These recordings vary widely in audio quality, airport environments, and speaker accents, capturing the realities of high-stakes aviation dialogue. 
While the Airbus-ATC corpus is mainly French accented, the ATCO2 corpus contains several accents, mainly Czech, Swiss German, and also Swiss French and Australian English.

In the context of ATC, messages can be categorized into three main types. 
First, communications from the air traffic controller serve as authoritative instructions or clearances directed to pilots.
Second, pilot transmissions function as responses or requests for clarification, position reports, or emergency declarations, facilitating coordination with air traffic controllers and ensuring adherence to given instructions. 
These two kinds of messages, from control agents and pilots, are present in both the Airbus-ATC and ATCO2 corpora. 
Finally, the Automatic Terminal Information Service (ATIS) provides continuous updates on meteorological conditions, runway availability, and operational notices relevant to a specific airport. 
ATIS is characterised by utterances
of about 30s in average, longer than regular exchanges between pilot and the ATC, which have an average duration of 4.5s.
ATIS messages are present in Airbus-ATC, but not in the ATCO2 corpus. 



In our study, we used 
around 4,500 hours of ATCO2 unlabelled audio recordings in English for SSL pre-training,
the official distribution of the Airbus-ATC 2018 challenge for ASR supervised fine-tuning (40h) and evaluation, the freely available ATCO2-test-1h\footnote{\url{https://huggingface.co/datasets/Jzuluaga/atco2_corpus_1h}} and the licensed ATCO2-test-4h for evaluation only.

\section{In-domain and out-domain SSL models}

\subsection{Self-supervised learning of in-domain models}
\label{sec:in-domain_ssl}
To pre-train in-domain models via self-supervision, we selected the BEST-RQ framework. 
This decision was influenced by its open-source availability within the Speechbrain project~\cite{ravanelli2024open} and its efficiency—2.5 times faster than wav2vec 2.0~\cite{whetten2024open}. 
In addition, it demonstrates performance comparable to the widely used wav2vec 2.0 approach~\cite{chiu2022self,whetten2024open}.
BEST-RQ is a self-supervised learning approach that uses a random-projection quantizer to turn speech signals into discrete labels, then trains a speech encoder to predict those labels for masked parts of the input.
Because the quantizer is fixed and untrained, it places fewer constraints on the encoder architecture—allowing both streaming and non-streaming models—and avoids the added complexity of jointly learning a representation~\cite{chiu2022self}.

For pre-training, we rely on the SpeechBrain recipe\footnote{\url{https://github.com/speechbrain/speechbrain/tree/develop/recipes/LibriSpeech/self-supervised-learning/BEST-RQ}} applied to 4,500 hours of unlabelled ATCO2 English audio. 
We trained a \textit{Large} model of 300M parameters with $848$ dimensions for hidden representations and $24$ encoder layers.
This model is pre-trained for 300K iterations by employing sixteen H100 GPUs. 
We select the batch size to optimize GPU memory usage, resulting in 2 hours of audio per batch. 
Training this \textit{Large} BEST-RQ model for 300K iterations requires approximately two days.
The masking strategy uses segments of four frames with a probability of 0.15, meaning that 15\% of segments are masked (i.e 60\% of speech).

This model is called \offlinemodel{Large}.

\subsection{Out-domain existing SSL models}

In order to compare our in-domain SSL models to existing out-domain models on the ASR task applied to ATC data, we made some experiments (described in section~\ref{sec:offline_ft}) by using several of the most popular ones: wav2vec2.0 models (XLSR-128~\cite{babu2021xls} and LS960~\cite{baevski2020wav2vec}), MMS-1B~\cite{pratap2023mms}, wavLM~\cite{chen2022wavlm}, HuBERT~\cite{hsu2021hubert}, and w2v-BERT~2.0~\cite{chung2021w2v,barrault2023seamless}.
In this paper, we focus on the two speech encoders that delivered the best results on our preliminary experiments as the other models had a significantly higher word error rate (WER). 
The two models we kept for this paper are HuBERT~\cite{hsu2021hubert} and w2v-BERT~2.0~\cite{chung2021w2v,barrault2023seamless}.
The HuBERT Large model used in this study has been pre-trained on 60,000 hours of unlabelled English speech from the Libri-Light dataset while w2v-BERT~2.0 has been pre-trained on 4.5 million hours of speech in 143 languages from diverse public datasets. 



\begin{table*}[h!]
\centering
\footnotesize  
\setlength{\tabcolsep}{4pt}  
\caption{ASR results of the two best out-domain speech encoders compared to our in-domain speech encoder on the Airbus-ATC dev and test corpora, and on the two ATCO2 test corpora.}
\begin{tabular}{|l|c|c|c|c|c|c|c|}
\hline
 SSL model & Pre-train & \#Par. & LM & A-dev & A-test & AT2-1h & AT2-4h \\ \hline
\hline          
w2v-BERT~2.0 & 4.5M & 600M & 4-gram & \textbf{6.74} & \textbf{6.21} & 23.12 & 29.30 \\ \hline
HuBERT & 60k & 300M & 4-gram & 7.70 & 7.25 & 33.33 & 39.26 \\ \hline
\offlinemodel{Large} & 4.5k & 300M & 4-gram & 7.90 & 7.40 & \textbf{19.70} & \textbf{28.70} \\ \hline
\hline
w2v-BERT~2.0 & 4.5M & 600M & -- & \textbf{7.36} & \textbf{7.01} & 25.98 & 31.74 \\ \hline
HuBERT & 60k & 300M & -- & 8.93 & 8.50 & 37.55 & 43.39 \\ \hline
\offlinemodel{Large} & 4.5k & 300M & -- & 10.67 & 10.01 & \textbf{24.27} & \textbf{30.73} \\ \hline
\end{tabular}
\label{table:speech_encoders}
\end{table*}

\section{Offline ASR on the ATC data}
\label{sec:offline_ft}

To compare the in-domain \offlinemodel{Large} model described in section~\ref{sec:in-domain_ssl} with popular out-domain SSL models, we fine-tune these models for ASR applied to ATC data.
As mentioned in section~\ref{sec:datasets}, we use the 40 hours of the labelled Airbus-ATC training data for these fine-tunings, in addition to the 5  hours for development purposes. 

\subsection{Offline ASR fine-tuning setup}
\label{sec:offline_ft_setup}
For the fine-tuning phase, we adopt as a downstream probe a straightforward architecture consisting of a 3-layers DNN followed by a linear layer and a softmax activation function. 
The training is performed using the Connectionist Temporal Classification (CTC) loss. 
The probe's hidden layers have a dimension of 1024, with a dropout rate of 0.15.

We use different learning rates and optimizers for the pre-trained encoder and the probe. 
We fine-tune the BEST-RQ encoder using a learning rate of $10^{-4}$ while the probe is trained with a higher learning rate of $8\times10^{-4}$.
w2v-BERT is fine-tuned using a learning rate of $1\times10^{-5}$ with a probe learning rate of $1.5$. 
HuBERT is fine-tuned using a learning rate of  $1\times10^{-4}$with a probe learning rate of $1.0$. 
The batch size is adjusted according to the encoder size to fit within an A100 80GB GPU, resulting in approximately 450 seconds of audio per batch for \textit{Large} BEST-RQ and HuBERT models and 40 seconds for w2v-BERT.
We fine-tune the entire model for 30 epochs with BEST-RQ and 80 epochs with HuBERT and w2v-BERT on Airbus-ATC training set, selecting the best checkpoint based on the WER obtained on Airbus-ATC development set.
A 4-gram language model (LM) was trained on the Airbus-ATC training data, ATCOSIM \cite{hofbauer2008atcosim} and UWB\_ATCC \cite{vsmidl2019air} datasets. 
We report on results with and without this 4-gram LM. When the LM is used we use beam search decoding with a beam size of 1000. When no LM is applied, the model defaults to greedy decoding.

\subsection{Experimental results}

Table~\ref{table:speech_encoders} presents the word error rate (WER) obtained on the Airbus-ATC and ATCO2 test sets by using the two existing out-domain SSL models we selected in regards with their performance and our in-domain SSL model, with and without language model integration.
w2v-BERT~2.0, with 600M parameters and significant pre-training on 4.5 millions hours of multilingual speech, demonstrates superior performance on Airbus datasets.
It achieves a WER of $6.21$\% on the Airbus-ATC test set and $23.12$\% on the ATCO2-1h subset when using a 4-gram language model. 
The HuBERT model, with its smaller architecture of 300M parameters pre-trained on 60k hours of English speech, shows competitive but slightly inferior performance, achieving $7.25$\% and $33.33$\% WER on the same subsets respectively.
Interestingly, \offlinemodel{Large}, despite its architecture of 300M parameters and limited pre-training on 4.5k hours of ATCO2 unlabelled data, achieves the best performance on the ATCO2-1h and ATCO2-4h subsets with $19.7$\% and $8.7$\% WER, significantly outperforming both w2v-BERT~2.0 and HuBERT. 
While both Airbus-ATC and ATCO2 datasets contain air traffic control communications, they present distinct acoustic characteristics. 
The ATCO2 corpus was collected through a network of very-high frequency (VHF) radio receivers operated by volunteers, resulting in specific acoustic conditions influenced by factors such as equipment quality (various antenna types and SDR receivers), signal reception, and environmental variables. Additionally,as mentioned in section~\ref{sec:datasets}, Airbus-ATC and ATCO2 do not contain the same accents.
\offlinemodel{Large}, being pre-trained specifically on this type of data, demonstrates particularly strong performance on the ATCO2 corpus, highlighting the importance of domain-specific pre-training for handling specialized acoustic conditions.

We observe that the integration of a language model consistently improves performances across all models.
This improvement is even more pronounced for \offlinemodel{Large}, particularly on the ATCO2-1h corpus, where the WER decreases by $4.57$ points (from $24.27$\% to $19.70$\%). 
These results suggest that while extensive pre-training can be beneficial for robustness, as demonstrated by w2v-BERT~2.0 performance on Airbus data, our in-domain pre-training approach with BEST-RQ is effective for specific ATC contexts, despite using significantly fewer hours.

\section{Streaming ASR for ATC}

\subsection{Streaming self-supervised learning}
\label{sec:streaming_ssl}
To implement a streamable version of BEST-RQ, we replace classical attention mechanisms in the Conformer blocks by chunked attention~\cite{zhang2020unified}. 
This approach divides the input sequence into chunks that group a given amount of frames. 
Within each chunk, frames can attend to all other frames in the same chunk. 
Additionally, chunks can attend to a limited number of previous chunks.  
We also integrate Dynamic Chunk Convolutions (DCConv)~\cite{li2023dynamic} instead of conventional convolution layers in the Conformer blocks and reuse the same chunk boundaries we used for chunked attention. 
Unlike conventional convolutions, which create a mismatch between training and inference due to access to future context beyond chunk boundaries, DCConv restricts the convolution operation to within-chunk frames.

With the purpose of developing a model that can flexibly adapt to different streaming requirements at inference time, we implement a mixed training strategy: 40\% of batches are trained with full context, without any chunking constraints, and 60\% of batches use dynamic chunking. Chunk size is randomly sampled between 8 and 32 frames, and 75\% of chunks have restricted left context (2-32 chunks), while 25\% maintain full left context.
This mixed training approach results in a model that can operate across different latency requirements at inference time, from low-latency streaming to full-context processing.

We train two BEST-RQ models with our streaming approach: a $Large$ version with the same size as the one trained in classical offline mode, called \streamingmodel{Large}, and a $Base$ one with 92M parameters, called \streamingmodel{Base}.
To move from $Large$ to $Base$, we decrease the dimension of  hidden representations from $848$ to $576$ and reduce the number of encoder layers from $24$ to $12$. 
Both models are pre-trained with 300k iterations. The $Base$ model is trained on four H100 GPUs with 1.6 hours of audio per batch while the $Large$ keeps the same settings as for its offline version.

\subsection{Streaming ASR fine-tuning setup}
For the fine-tuning phase in streaming mode, we reuse the probe architecture described in section~\ref{sec:offline_ft}. 
We modify the supervised learning strategy by applying dynamic chunking to 100\% of the batches, rather than the mixed approach used during the SSL pre-training.

\subsection{Experimental results on streaming ASR}

\begin{figure*}[!t]
    \centering
    \includegraphics[width=1.0\textwidth]{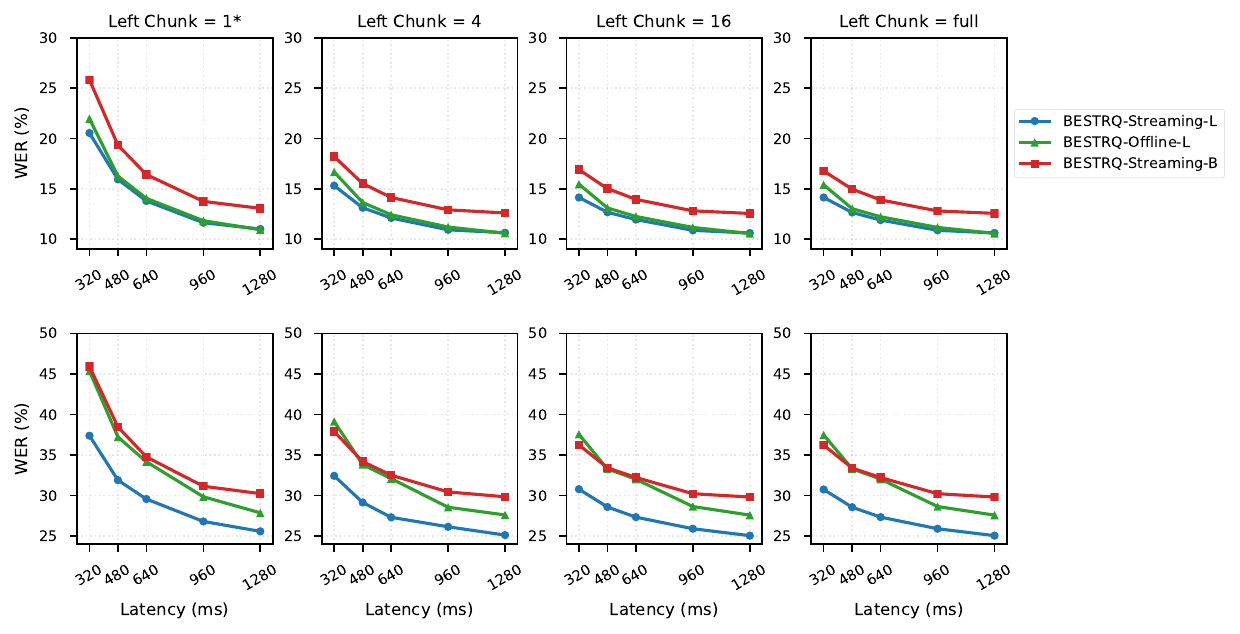}
    \caption{Performances of BEST-RQ encoders fine-tuned on the \textbf{streaming} ASR task. (top) Results on the Airbus-ATC test data. (bottom) Results on the ATC02-1h test data. Different left contexts and (right) chunk sizes are investigated.}
    \label{fig:streaming}
\end{figure*}

The streaming capabilities of our BEST-RQ streaming models on both Airbus-ATC and ATCO2 test sets are shown in Figure~\ref{fig:streaming}. We compare three variants of SSL models: \streamingmodel{Large}, \offlinemodel{Large} (pre-trained without streaming strategy), and \streamingmodel{Base} models.
Figure~\ref{fig:streaming} presents the WER results across different left context sizes and chunk sizes, which correspond to controlled model's latency. No language model was integrated.
On the Airbus-ATC test set, \streamingmodel{Large} achieves the best performance across all configurations, ranging from $10.6$\% to $20.55$\% depending on streaming settings.
With full left context, it reaches a WER of $10.6$\%  at a 1280ms latency, showing minimal degradation compared to the offline model ($10.6$\%).
With aggressive streaming constraints (left chunk = 1), which has not been seen during training, the model achieves $20.55$\% WER at 320ms of latency, demonstrating the robustness of the model in low-latency scenarios. 
The impact of streaming adaptation is evident on the ATCO2 dataset, where the streaming fine-tuning shows substantial improvements.
The \streamingmodel{Large} model outperforms its offline counterpart by a fair margin, achieving $25.07$\% WER versus $27.6$\% WER at 1280ms of latency with full left context.
The performance gap between the two models is even bigger at lower latencies. 
For both datasets, larger left context sizes improve performance, but improvements become minimal beyond 16 chunks. 
Finally, the \streamingmodel{Base} model shows competitive performance, particularly on the ATCO2 dataset when compared to the offline \offlinemodel{Large} model fine-tuned for streaming ASR. 
These result highlight the viability of the \textit{Base} architecture for applications where computational resources are limited, and show that our streaming SSL and fine-tuning strategies effectively adapt the BEST-RQ models for real-time ASR applications for ATC data, particularly  for strict latency constraints.

\subsection{SSL streaming models applied to offline ASR}

We also tested the performance of our models pre-trained through our streaming SSL approach but used for ASR in offline mode, without latency constraints. In this case, we fine-tuned the models following the offline mode described in section~\ref{sec:offline_ft_setup}.

Despite restricting the context in pre-training, we can observe that the $Large$ model pre-trained in streaming mode outperforms in WER the offline \offlinemodel{Large} model pre-trained in a conventional way in all the test datasets (Table~\ref{tab:offline_wer_comparison}), even outperforming the HuBERT model (see Table~\ref{table:speech_encoders}).




\begin{table}[htbp!]
\centering


\caption{WER of the different BEST-RQ models and the Airbus-ATC and ATCO2 test datasets for \textbf{offline} ASR with the a 4-gram language model. Pre-training in a streaming fashion by restricting context, proved to be helpful even in an \textbf{offline} (non-streaming) ASR setting.}

\begin{tabular}{|l|c|c|c|}
\hline

SSL model & Airbus & ATC02-1h & ATC02-4h  \\
\hline
\hline
\offlinemodel{Large} & \textcolor{black}{7.40} & \textcolor{black}{19.70} & \textcolor{black}{28.70} \\
\hline

\streamingmodel{Large} &  \textcolor{black}{\textbf{7.18}} & \textcolor{black}{\textbf{19.30}} & \textcolor{black}{\textbf{26.59}}\\
\hline

\streamingmodel{Base} & \textcolor{black}{8.09} & \textcolor{black}{24.58} &  \textcolor{black}{29.47}\\
\hline

\end{tabular}
\label{tab:offline_wer_comparison}
\end{table}

To further analyze the  performance of the streaming SSL model on the offline ASR task, we computed the WER based on the type of messages (see section~\ref{sec:datasets}).
Table~\ref{tab:wer-speaker-role} presents the WER for controller, pilot, and ATIS messages in the Airbus-ATC test corpus.
An improvement is observed for each type of message, with the smallest occurring in ATIS messages (-1.37\% relative) and the largest in the noisiest category, pilot messages (-3.46\% relative).
These results suggest that the mixed SSL training approach using both full context samples and dynamic chunking is particularly useful to process noisy recordings.

\begin{table}[h!]
\centering
\caption{WER according to speaker role (C:controller, P:pilot, A:ATIS) on the Airbus-ATC test corpus for \textbf{offline} ASR with a 4-gram language model.}
\begin{tabular}{|l|lll|}
\hline
 SSL model   & \multicolumn{1}{l|}{C} & \multicolumn{1}{l|}{P} & A  \\ \hline
\hline
\offlinemodel{Large}  & \multicolumn{1}{l|}{\textcolor{black}{4.84}}  & \multicolumn{1}{l|}{\textcolor{black}{10.41}}  &   \textcolor{black}{5.84}  \\ \hline
\streamingmodel{Large}  & \multicolumn{1}{l|}{\textcolor{black}{\textbf{4.73}}}  & \multicolumn{1}{l|}{\textcolor{black}{\textbf{10.05}}}  &   \textcolor{black}{\textbf{5.76}}  \\ \hline

\end{tabular}
\label{tab:wer-speaker-role}
\end{table}
\section{Discussion}

In this work, we investigated how in-domain SSL pre-training impacts both offline and streaming ASR for Air Traffic Control (ATC) communications. 
We found that while large-scale, general-purpose models (e.g., w2v-BERT 2.0 with 4.5M hours) excel at broad tasks, our BEST-RQ model—which uses only 4.5k hours of domain-specific data—outperforms them on the ATCO2 corpus. This shows that targeted in-domain pre-training can be more effective than large-scale general-purpose training for the unique acoustic conditions of VHF radio communications.
We also demonstrated that our streaming strategy---combining chunked attention, dynamic convolutions, and a mixed training approach using both full context samples and dynamic chunking---is highly effective for real-time ATC speech recognition. 
Notably, this approach also proves beneficial for offline ASR.
Our BEST-RQ Large model pre-trained in streaming mode retains strong performance across different latency settings, with only minimal degradation compared to its offline counterpart. 
Even under strict latency limits, the model remains practical.
This model also achieves better offline ASR performance than its counterpart pretrained using a conventional offline approach.
Overall, these results highlight the value of specialized SSL pre-training combined with a streaming approach for focused speech recognition tasks. 
Future work could involve integrating these models into operational ATC systems and examining their robustness to the diverse accents and noise conditions common in ATC communications.

\newpage

\begin{credits}
\subsubsection{\ackname} This work used HPC resources from GENCI-IDRIS: grants AD011012551R3, AD011015051R1, AD011012108R3, AD011014814R1, and AD011015509.

\end{credits}

%
%
%
\bibliographystyle{splncs04}
\bibliography{mybib}

\end{document}